\documentclass[11pt]{article}
\usepackage{amsmath, amssymb, amsthm}
\usepackage{graphicx}
\usepackage{geometry}
\usepackage{placeins}
\usepackage{titlesec}
\usepackage{hyperref}
\usepackage{tabularx}

\usepackage{mathpazo}

\title{\textbf{Scalable Multi-Robot Path Planning via Quadratic Unconstrained Binary Optimization}}
\author{Javier González Villasmil}
\date{\today}

\begin{document}
\maketitle

\begin{abstract}

Multi-Agent Path Finding (MAPF) remains a fundamental challenge in robotics, where classical centralized approaches exhibit exponential growth in joint-state complexity  as the number of agents increases. This paper investigates Quadratic Unconstrained Binary Optimization (QUBO) as a structurally scalable alternative for simultaneous multi-robot path planning. This approach is a robotics-oriented QUBO formulation incorporating BFS-based logical pre-processing (achieving over 95\% variable reduction), adaptive penalty design for collision and constraint enforcement, and a time-windowed decomposition strategy that enables execution within current hardware limitations. An experimental evaluation in grid environments with up to four robots demonstrated near-optimal solutions in dense scenarios and favorable scaling behavior compared to sequential classical planning. These results establish a practical and reproducible baseline for future quantum and quantum-inspired multi-robot coordinations.

\end{abstract}

\section{Introduction}

Swarm robotics and autonomous multi-agent navigation are rapidly advancing and compelling areas of robotics research. Coordinated multi-robot systems offer significant operational advantages and a broader spectrum of capabilities over single agents but introduce substantial coordination and planning complexities. Multi-robot systems are increasingly deployed in warehouse automation, inspection, and distributed sensing tasks, wherein multiple agents must navigate shared environments efficiently and safely. A fundamental challenge in these systems is Multi-Agent Path Finding (MAPF), which involves computing collision-free paths for multiple robots while optimizing performance metrics such as travel time or distance.\\
\\
Despite significant advances in classical MAPF algorithms, centralized formulations suffer from combinatorial growth in the joint-state complexity. As robotic swarms grow in size and coordination requirements intensify, alternative formulations that better manage the combinatorial complexity become increasingly relevant.\\
\\
This study investigates Quadratic Unconstrained Binary Optimization (QUBO) as a unified optimization framework for simultaneous multi-robot path planning. Unlike sequential approaches, QUBO encodes all agents within a single global optimization problem, yielding a formulation whose variable count scales linearly with the number of robots while maintaining an NP-hard structure.\\
\\
To make QUBO practical for robotics applications, we introduced the following:

\begin{itemize}

\item A robotics-specific QUBO encoding for MAPF with structured penalty design,
\item A BFS-based logical pre-processing strategy achieving over 95\% variable reduction,
\item A time-windowed decomposition method enabling execution within the constraints of current quantum and quantum-inspired hardware,
\item An empirical evaluation establishing baseline performance relative to classical planning.

\end{itemize}

The goal of this work is not to claim a quantum advantage at current hardware scales but to establish a reproducible and scalable formulation that bridges classical robotics and emerging quantum-inspired optimization.

\section{Related Work}

\subsection{Classical Multi-Agent Path Finding}

Multi-Agent Path Finding (MAPF) pursues collision-free paths for multiple agents in shared environments. Classical approaches include Conflict-Based Search (CBS) \cite{SHARON2015_CBS}, which plans individual paths and resolves conflicts hierarchically, and prioritized planning \cite{cap2014prioritizedplanningalgorithmstrajectory}, which assigns priorities to agents and plans sequentially. Although CBS is complete and optimal, it suffers from exponential worst-case complexity in the number of agents \cite{mapf}. Time-expanded graphs \cite{Yu_LaValle_2013} provide an alternative representation but face similar scalability challenges in dense multi-agent scenarios.

\subsection{QUBO and Quantum Optimization}

Quadratic Unconstrained Binary Optimization has been successfully applied to combinatorial problems, including Max-Cut \cite{rehfeldt2022fasterexactsolutionsparse}, Traveling Salesman Problem \cite{lucas2014ising}, and job scheduling \cite{venturelli2016job}. D-Wave Systems has demonstrated quantum annealing for traffic flow optimization \cite{neukart2017traffic} and vehicle routing \cite{Feld_2019}, though these focus on single-vehicle path optimization rather than multi-agent coordination.\\
\\
The Quantum Approximate Optimization Algorithm (QAOA) \cite{farhi2014quantumapproximateoptimizationalgorithm} provides a gate-based alternative to quantum annealing, with applications to graph problems \cite{Zhou_2020}. However, direct applications of this method to robotic path planning with collision constraints remain largely unexplored. Existing QUBO formulations for routing lack the penalty structures required for multi-agent coordination in grid environments.

\subsection{Gap This Work Addresses}

While classical MAPF algorithms perform well on small-scale problems and QUBO methods have demonstrated potential for combinatorial optimization, recent advancements in automated QUBO formulation for general pathfinding problems, such as the framework proposed by \cite{rovara2024frameworkformulatepathfindingproblems}, have begun to address encoding challenges for instances, including multi-agent scenarios. These approaches primarily emphasize constraint translation and automation, particularly for small, graph-based problems.\\
\\
However, they fall short of addressing robotics-specific requirements, such as comprehensive collision avoidance in grid environments, logical pre-processing for variable reduction, handling of solver failure modes, and decomposition strategies essential for practical deployment on current quantum hardware.\\
\\
This study builds on these foundations to bridge the remaining gaps in quantum-inspired multi-agent robotic path planning by (1) formulating MAPF as a QUBO with robot-specific penalties tailored for dynamic coordination, (2) introducing BFS-based pre-processing paired with a time-windowing approach that reduces variables by $  >95\%  $, enabling scalability beyond existing qubit limitations, and (3) providing an honest empirical comparison that establishes baseline performance for future quantum approaches.

\section{Optimization Background}

This study argues that QUBO-based optimization provides a structurally scalable alternative to classical sequential MAPF formulations, even in the absence of near-term quantum advantage. This section summarizes the optimization concepts necessary to understand the formulation. Readers familiar with QUBO and quantum-inspired optimization may skip up to Section 4.

\subsection{Quadratic Unconstrained Binary Optimization}

The mathematical framework we will follow is QUBO, which facilitates the natural modeling and solution of path planning problems. Although QUBO does not surpass classical algorithms such as A* or Dijkstra for single-agent scenarios, its structure provides specific advantages in multi-agent path planning. In dynamic environments, conventional methods typically address agents sequentially, resulting in sub-optimal outcomes.\\
\\
In contrast, QUBO formulates the system as a unified optimization problem, allowing simultaneous planning across all agents. This reduces the exponential joint-state complexity of the centralized MAPF to a formulation in which the variable count scales linearly with the number of agents, while the optimization problem remains NP-hard. This characteristic makes QUBO suitable for a server-based approach in multi-robot systems, although distributed solutions remain standard in robotics.

\subsubsection{Definition}

The Quadratic Unconstrained Binary Optimization (QUBO) model provides a unified mathematical framework for representing discrete optimization problems. It consists of minimizing a quadratic function defined over binary variables, which is typically expressed as
\begin{equation}
C(x)=x^TQx
\label{QUBO definition}
\end{equation}

where 
\( x\in\{0,1\}^n \) represents the decision variables and 
\( Q\in\mathbf{R}^{n\times n} \) encodes the relationships between them. It is common to assume that the matrix \(Q\) is symmetric or in the upper triangular form; in fact, most solvers use the upper triangular approach mixed with \textbf{sparse algebra} for efficiency.\\
\\
The model is unconstrained in the sense that all problem conditions are embedded directly into the cost function, rather than expressed as separate constraints. Its quadratic nature arises from the pairwise interactions between binary variables represented by the cross terms in 
\(Q\). This structure enables the QUBO to capture complex dependencies and penalty relationships in a compact form.\\
\\
This nature is flexible enough to handle multiple objectives, which is important for path planning, particularly in multi-agent scenarios. Because it is unconstrained, the user defines their own constraints (i.e., there is no unique QUBO structure for a topic). It encodes both the objectives and penalties into a single cost function. 
\[
F(x)= \lambda_1\cdot\text{distance\_penalty}(x)+\lambda_2\cdot\text{obstacle\_penalty}(x)
\]
This nature also makes it a standard for various optimization approaches. Many famous NP-hard problems, such as Max-Cut, Max-SAT, and Knapsack, can be encoded into the QUBO form. Its ability to encompass many models in combinatorial optimization is enhanced by the fact that the QUBO model is equivalent to the \textbf{Ising model}, which plays a prominent role in physics.
Note that QUBO is not directly solved; it solely expresses problems naturally. Then, it can be converted into a Hamiltonian to be solved using quantum Annealing or QAOA. It can also be converted into an Ising model to follow a physics representation, where instead of using variables \(\ \{0,1\}\) it uses \( \{+1,-1\} \) respectively.

\subsubsection{Properties of QUBO}

The most important property of a QUBO is that it is quadratic, which means that it can be multiplied by at most two variables. Although these properties can be bypassed by easily introducing extra variables to only do a quadratic term (the price to pay would be that one extra variable would be equivalent to the quadratic term of the other, so that the quadratic term can be chained with other variables and have a theoretically higher degree of association rather than a basic quadratic.
Another interesting fact is that we can classify the equation \( C(x)\) (\ref{QUBO definition}) into linear and quadratic terms. Let us suppose the following equation:  \(C(x)=-5x_1-3x_2-8x_3-6x_4+4x_1x_2+8x_1x_3+2x_2x_3+10x_3x_4\)
. The linear part of the equation is \(-5x_1-3x_2-8x_3-6x_4\) while the quadratic part is \(4x_1x_2+8x_1x_3+2x_2x_3+10x_3x_4\). However, since binary variables satisfy \(x_j=x^2_j\), the linear part can also be written as quadratic \(-5x_1^2-3x_2^2-8x_3^2-6x_4^2\). In particular, if we follow the formal definition of  (\ref{QUBO definition}), it can be written as \[
C(x)=\begin{bmatrix}x_1 & x_2 & x_3 & x_4 \end{bmatrix} \begin{bmatrix} -5 & 2 & 4 & 0 \\ 2 & -3 & 1 & 0 \\ 4 & 1 & -8 & 5 \\ 0 & 0 & 5 & 6 \end{bmatrix} \begin{bmatrix} x_1 \\ x_2 \\ x_3 \\ x_4 \end{bmatrix}
\]
It can be observed that the linear terms determine the elements on the main diagonal of \(Q\) and the quadratic terms determine the off-diagonal elements. This will be studied more deeply in the pre-processing part (which will help greatly reduce the solution times). The diagonal terms represent the individual weights of each variable, whereas the off-diagonal terms represent the weights between different variables.

\subsubsection{Hamiltonian}

The Hamiltonian encodes the energy (or cost) of a particular binary configuration, similar to how a QUBO function assigns a "score" to each binary string. The ground state of the Hamiltonian is the answer that minimizes the QUBO and hence the optimal solution to the optimization problem. To solve the QUBO, we look for the eigenvalue and eigenvector of its corresponding Hamiltonian, where the eigenvalue represents the energy, and the eigenvector represents the solution string.

\subsubsection{Why is QUBO suited for path planning?}

QUBO is an adequate representation of the Path Planning problem because most interactions are inherently quadratic. Real-world routing involves what we call "interdependent constraints" - factors that interact with each other in non-linear ways. For example, taking a slightly longer route might save fuel if it avoids stop-and-go traffic, but this relationship is quadratic because it depends on both the route choice and traffic conditions simultaneously.\\
\\
Consider a delivery truck planning multiple stops; the optimal route depends not only on the distances between points but also on how the weight distribution changes as packages are delivered, how traffic patterns vary throughout the day, and how different route choices interact with vehicle constraints. These are naturally quadratic relationships that linear algorithms struggle to capture.\\
\\
It is also particularly good for multi-agent path planning because collision avoidance between agents must be handled, which creates natural quadratic interaction terms between different agents' path variables. We can also benefit from the weight of the penalties, which allows us to make the QUBO more dynamic. The \(\lambda\) values can be a function of the problem parameters. For a delivery truck, \( \lambda_{fuel}\) might increase as fuel prices rise, or \(\lambda_{time}\) might increase during peak traffic hours (making more efficient de-routes that avoid traffic).

\subsubsection{Sparse Algebra}

Sparse algebra problems occur when matrices have a significantly large number of zeros; therefore, there are benefits from explicitly considering zeros when solving the problems.\\
\\
In this case, these zeros arise from the fact that the matrix is symmetric, which means that from the pure start, we can only represent it by using the upper triangular half without explicitly writing the other half. Second, it consists of a matrix where off-diagonal elements represent the relation between two variables; however, it is most likely that not all the variables are related, rather than for them to actually contain values. It benefits from the symmetrical structure and the fact that it tends to have many zeros. This helps reduce both memory usage and computation time.\\
\\
This is the type of matrix we use to represent QUBOs, and thanks to that, pre-processing helps reduce the computation time and memory.

\FloatBarrier
\begin{figure}[ht]
\centering
\begin{minipage}{0.45\textwidth}
  \centering
  \includegraphics[width=\linewidth]{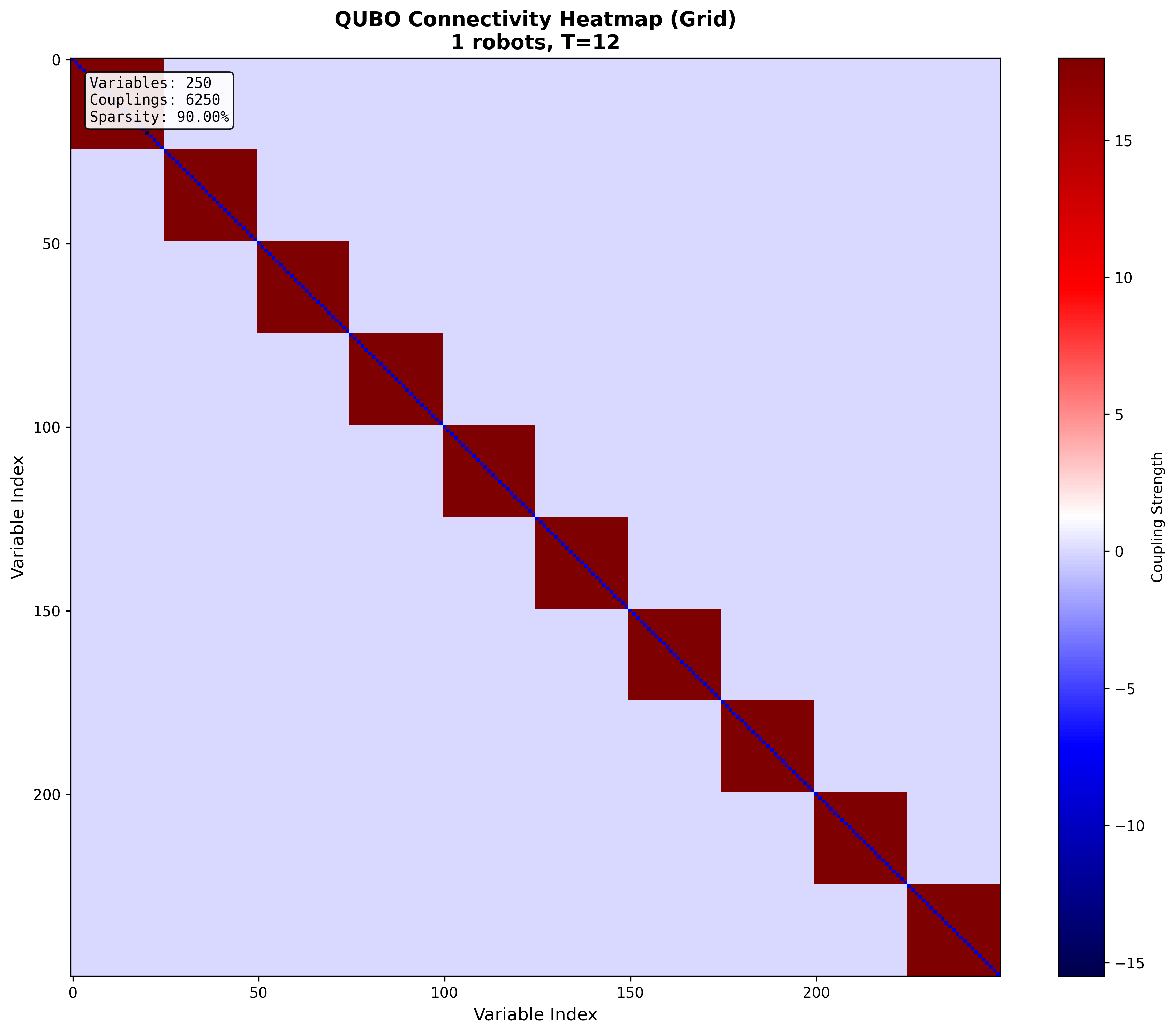}
  \caption{Sparse matrix visualization of a single-robot QUBO formulation on a 5×5 grid. Dark regions indicate non-zero entries representing penalty interactions.}
\end{minipage}\hfill
\begin{minipage}{0.45\textwidth}
  \centering
  \includegraphics[width=\linewidth]{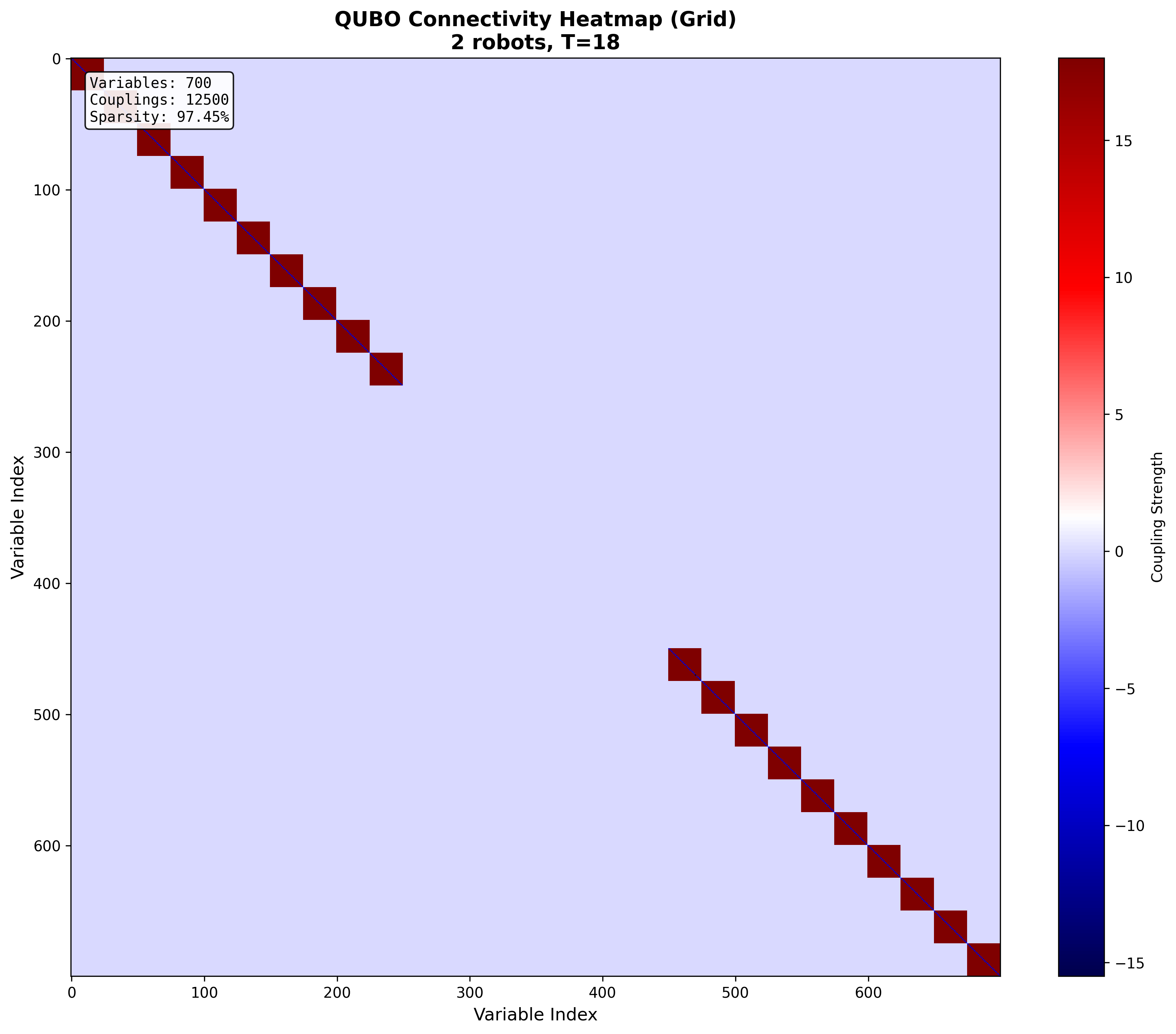}
  \caption{Sparse matrix for two-robot scenario showing increased sparsity. Note the separation corresponding to each robot's variable space.}
\end{minipage}
\end{figure}
\FloatBarrier

\subsection{Quantum Annealing}

A quantum annealer is a type of quantum computer designed to solve optimization problems. These quantum computers are not meant to be freely programmed by the user; they are built to support only a single quantum algorithm, which is \textbf{quantum annealing}. Although this continuous approach can be more natural for certain problems, it may lack the universality of gate-based quantum computing. \\
\\
Since we are formulating the optimization problem as a QUBO, it can be easily solved via quantum annealing; in this case, we only need to care about problem specification and not about solver characteristics or parameters. However, we want to set an ideal number of samples and normalization scale for the QUBO that lets us find the optimal solution in a consistent manner in a good time.\\
\\
The algorithm is based on the \textbf{Adiabatic Theorem}, which is a method to find the global minimum of the objective function without checking all combinations or local minima. First, we define two Hamiltonians: the initial and final Hamiltonians. The initial Hamiltonian is a simple objective function for which the global minimum is known, whereas the final Hamiltonian is the optimization problem that we are trying to solve. Remember that a QUBO can be easily converted into an Ising Hamiltonian with a single conversion equation. The idea is to gradually modify the shape of the initial Hamiltonian into an equivalent objective function of the optimization problem. If the annealing process happens slowly enough without disturbances, the theorem ensures that the global minimum point has adapted to the shape of the function in all the transformation phases of the objective function. \cite{Kadowaki_1998}\\
\\
Nowadays, there is only one company dedicated to the design of quantum annealing computers, D-Wave\cite{DwaveQA}. They use qubits whose states are defined by a circulating current and a corresponding magnetic field. A magnetic field is applied programmatically via a qubit bias to push the probabilities into one state or the other. They shape qubit probabilities to be of the form of the optimal solution. The qubits can be linked together so that they can influence each other, and D-Wave does so via \textbf{couplers}, devices that correlate and entangle qubits, which affect the joint probabilities and in the case of entanglement, a quantum phenomenon that with a partial view of the qubits (environment), you can predict the other qubit (outcome) with full certainty; although this a topic to be treated more deeply and separately for quantum robotics, especially in the area of \textbf{swarm robots} and \textbf{coordinated movements}.\\
\\
It is important to know, however, that although the adiabatic theorem states that the optimal solution should always be found, external noise can cause the system to not reach that local minimum of the final Hamiltonian. The most common issue is \textbf{thermal fluctuations}, which affect the system and make it nearly impossible to have a fully isolated system. For this reason, quantum computers must run in extremely low temperatures; particularly, D-Wave quantum annealing computers run around -273 degrees Celsius. Another problem is the rhythm of change of the initial function; if the annealing is too slow, then the environment may start to cause interference; however, if it is too fast, it gets stuck in local minima.\\
\\
In summary, it is an algorithm to take into account for its easy implementation, since you do not need to handle the underlying logic and parameters of the quantum computer, and because it is accessible via D-Wave's platform that connects you to real quantum annealing hardware.

\subsection{Quantum Approximation Optimization Algorithm}

The Quantum Approximation Optimization Algorithm (QAOA) is a variational quantum algorithm that alternates between a "problem Hamiltonian" that encodes the optimization problems and a "mixer Hamiltonian" that explores different solution states \cite{farhi2014quantumapproximateoptimizationalgorithm}. It is a hybrid quantum-classical algorithm that combines quantum circuits and classical optimization to find approximate solutions to combinatorial optimization problems. Conversely, quantum annealing is an algorithm based on quantum-gate system evolution.\\
\\
The algorithm is described by two main components: a parametric quantum circuit and classical optimization. The parametric quantum circuit, also known as an ansatz circuit, consists of gates that encode the optimization problem to be solved, where these gates are determined by parameters that need to be optimized. The evaluation of the system gives the most likely state, which is a potential optimal solution. The classical optimization part uses classical algorithms to optimize the parameters of these quantum gates. The process is iterative and involves reapplying the quantum gates using the latest set of optimized parameters until a near-optimal solution is found \cite{QAOA}.\\
\\
The problem (or cost) Hamiltonian encodes the optimization problem we want to solve. It is constructed by first formulating the optimization problem as a QUBO, then transforming it into an Ising Hamiltonian using the following change of variables: \(x_1=\frac{1-z_1}2\). The mixer Hamiltonian introduces dynamism to the system and allows the transition between states. With the mixer Hamiltonian, one can explore the eigenspectrum of the Hamiltonian and reach those good eigenstates. The most common mixer is the sum of Pauli X-Gates applied to each qubit of the circuit, which means changing the qubit state from 0 to 1 or vice-versa. However, you can use more complex mixers that are more adequate to the specific problem and can give better results. In my case, I will initially follow the simple mixer Hamiltonian while setting all penalties for path planning in the cost Hamiltonian. The mixer will be allowed to explore invalid states in this way, but the optimizer learns to avoid them because the cost Hamiltonian makes invalid solutions energetically expensive. Although these invalid states can also be handled with QUBO pre-processing, they will be shown in the following sections. Another important thing to know is that all qubits of the ansatz circuit should be started on the superposition state (Hadamard Gate). The point is that initially, there is only one global minimum, and there is no differentiation or bias toward any solution at the start (this is related to the idea of the initial Hamiltonian from the adiabatic theorem).\\
\\
Once the best parameters for the ansatz circuit have been determined, the mixer Hamiltonian must be applied to obtain that optimal solution in a very short time. It is important to know that in real life, you will not look for the specific perfect parameters for each path-planning problem; however, some general good parameters are determined, and depending on problem specifications, some noise is added to converge to the optimal solution.\\
\\
As can be seen, this algorithm offers several points of flexibility. Users can select their own ansatz circuit tailored to the optimization problem. The critical choices then become the initial parameters and classical optimizer, which together determine how effectively the algorithm converges to a near-optimal solution. As with other probabilistic algorithms, there's a tradeoff to manage: using enough samples to reliably obtain the optimal answer while minimizing computational overhead.

\begin{figure}[htbp]
    \centering
    \includegraphics[width=0.9\textwidth]{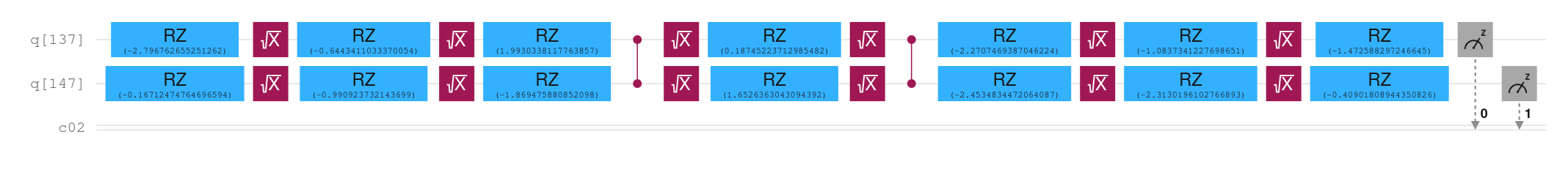}
    \caption{Example QAOA circuit for a small pathfinding problem with 2 qubits (after pre-processing). The circuit alternates between cost and mixer Hamiltonians for p=2 layers.}
    \label{fig:QAOA_circuit_2x2}
\end{figure}

\section{Path Planning}

Now that we have defined the necessary background for the quantum and mathematical concepts, we can focus on the path planning area.

\subsection{QUBO Implementation}

Before implementing QUBO, the first thing to understand is that the approaches we will be using to solve QUBO are not \textbf{sound} or \textbf{complete} algorithms. That is, they may not find a valid path if one exists, and it may return invalid paths. This could happen because either the QUBO was poorly formulated or the adiabatic theorem was broken, and it would suffer disturbances when trying to find the ground state. However, if the QUBO is well formulated, the eigenvector corresponding to the minimum eigenvalue of the diagonalized Hamiltonian should always be the optimal solution. Therefore, it is a semi-complete algorithm that, with enough runs, should have a very low probability of not finding the answer.\\
\\
Taking that into consideration, we want to formulate the QUBO in such a way that we reduce/penalize the bad paths and reward the good paths. This way, we reduce the chances of not finding the path and reduce the number of runs to find the good/optimal paths.\\
\\
The first thing we need to decide is the encoding of the QUBO. We already saw that it returns binary variables with their values set to either 1 or 0, and we have to interpret that sequence of binary variables. The most straightforward encoding would be assigning each position at time T a binary variable; if the variable is set to 1, then it is a position used for the path; otherwise, that position means nothing. Therefore, for the grid map structure, that would translate into MxNxT variables needed to encode the path planning problem, where MxN would be the matrix representation of the grid (M rows and N columns). For node graph representation, it would be NxT binary variables, where N is the number of nodes of the graph.\\
\\
So we know that MxNxT binary variables are sufficient to solve the optimization problem. However, we now need to define those variables, essentially, how we will encode the variable to decode a path given the bits. There are many options, and the user may choose whichever fits best or is considered better; however, we will follow the linear approach. 
\[
x_{i,j,t}\to x_n\\
n = i\cdot M + j + (MxN)\cdot T
\]
Basically, it groups variables per time step and enumerates the grid in a standard reading format, right to left, from top to bottom. In node graphs, it follows the node index (this can be custom or made to follow a grid representation). Understand that the encoding used is not trivial; it plays a key role when doing pre-processing.\\
\\
One could also say that the minimum number of binary variables required for the problem would be \(log2(MxN) x T\). Although theoretically correct, this does not mean that it is better. It would require a complex encoding, and that would disable the possibility of pre-processing (since now bits are no longer square independent, but would require one to treat them as a whole array of bits). Future work will explore this comparison; ultimately, if it makes it faster, I consider it to work best (although it would make logical pre-processing more challenging).

\subsubsection{Penalties}

Penalties will play a key role in the creation of the QUBO because they are responsible for crafting the desired solution. By nature, QUBOs are unconstrained; that is, the user is the one who defines their own constraints via penalties and which scenarios it wants to reward and which ones it wants to penalize (we consider a reward to be the same as a negative penalty, that is, one that minimizes energy).\\
\\
There are four essential penalties that each QUBO must follow for the answer to be valid. The main penalties are \textbf{one-hotness}, \textbf{adjacency}, \textbf{start condition}, \textbf{end condition}.\\
\\
\textbf{One-hotness} penalty ensures that the agent can only be present at one cell per time step. This is very important because in classical physics (not considering superposition since we are talking about large-enough objects), an object cannot be in two different places at the same time; its position must be unique. The math to encode this penalty is the following:
\[
\begin{aligned}
   P_{hot}&= K_{hot} \cdot \sum^T_{t=0}(1-\sum_{i,j}x_{i,j,t})^2 \\[1em]
   &(1-\sum x)^2 = (1-\sum x)(1-\sum x) = 1 - 2\sum x +
   (\sum x)^2 \\
&=1-2\sum x + (\sum x + \sum x_ix_j) \\ 
&= 1 - \sum x + 2 \sum x_ix_j \\
\left( \sum_{i=1}^{n} a_i \right)^2 &= \sum_{i=1}^{n} \sum_{j=1}^{n} a_i a_j = \sum_{i=1}^{n} a_i^2 + 2 \sum_{1 \leq i < j \leq n} a_i a_j
\end{aligned}
\]
\textbf{Adjacency} penalty ensures a continuous path. Its purpose is to avoid teleportation from one grid to another. This penalty can be set either for 4-connectivity (up, down, left, right) or 8-connectivity (where you can also move diagonally). However, when we refer to the adjacency, we will talk about the 4-connectivity scenario. This will be encoded as:
\[
P_{adj}=K_{adj} \cdot \sum_{t=0}^{T-1}\sum_{i,j}x_{i,j,t}(1-\sum_{(k,l)\in N(i,j)}x_{k,l,t+1})
\]
Adjacency is the most computationally expensive penalty among all the penalties that will be presented. It requires concatenating many loops. The \((k,l)\) are valid neighbors according to the current position. The best approach is probably to save them in a dictionary rather than calculating the neighbors each time on the fly. It is also important to realize that in this list, I did not include the cell itself; it just seems a waste of time to stay in place. In general, it is probably better to keep moving, although for multi-agent scenarios, it becomes necessary to add the ability to wait in place.\\
\\
The \textbf{start condition} ensures that QUBO returns at the first position of the path the correct start cell. It ensures the QUBO path starts at the intended block.
\[
\begin{aligned}
P_{start}&=K_{start} \cdot (1-x_{si,sj,0})^2 \\
&= K_{start} \cdot(1-2x_{si,sj,0}+x_{si,sj,0}^2) \\
P_{start} &= -K_{start} \cdot x_{si,sj,0}
\end{aligned}
\]
The \textbf{end condition} ensures that the QUBO reaches the goal. In practice, it will not always reach the goal, but this penalty is very important for the QUBO to look for the optimal path to the goal; without this, it would not know where to aim.
\[
P_{end}=K_{end}(1-x_{ei,ej,T})^2 = -K_{end} \cdot x_{ei,ej,T}
\]
There are two additional penalties that we could consider as principal too, since they are needed to measure if the solution is valid or not. These are: \textbf{Obstacle avoidance} and \textbf{Goal lock} constraints.\\
\\
\textbf{Obstacle avoidance} penalty is explanatory by itself, it ensures the agent does not crash against obstacles. It can be encoded as: 
\[
P_{obs}=K_{obs} \cdot\sum_{t=0}^T x_{i,j,t} \quad \forall \quad x_{i,j} \quad \text{obstacle} 
\]
However, this constraint can be handled indirectly. That means that you can ensure the robot does not crash into obstacles without explicitly writing a constraint. In the case of grid-map structures, you would merely need to remove the obstacles from the adjacency lists so that moving to an obstacle can be considered as breaking the adjacency penalty (which is a main constraint). In the case of nodes, it is even easier; obstacles are not even written in graphs, so adjacency encoding should also suffice, since it would make sure no connection between unconnected nodes. With these alternatives, obstacle avoidance is enforced structurally and not via penalty.\\
\\
The \textbf{goal lock} constraint looks for the agent to stay at the goal once it reaches it. I do not consider it a key penalty because you can solely consider the following moves after reaching the goal as extra and clip or ignore them. Also, this is some characteristic that standard algorithms do not worry about because they do not work with time-steps; however, the QUBO approach does not inherently know the right time window, so it is normal that it returns more time-steps than needed. Ideally, it stays there after reaching the goal. But notice that in practice, this penalty can be dropped without much problem.
\[
\begin{aligned}
P_{lock} &= K_{lock} \cdot \sum_{t=0}^{T-2}x_{g,t}(1-x_{g,t+1}) \\ &x_{g,t}(1-x_{g,t+1})  = x_{g,t} - x_{g,t}x_{g,t+1}
\end{aligned}
\]
\\
You probably noted that all formulas have a different constant attached to them. These constants weigh the importance of each constraint and play a crucial role. These were decided empirically; there does not seem to be a mathematically perfect weight for each penalty. When deciding penalty weights, you need to test the most general case (the one with fewer constraints) to make constraints global and not scenario-dependent. Otherwise, those constants may work only for constrained scenarios but not in the majority of cases, and this tends to happen because the space solution is larger and the solver has more freedom to ignore penalties.\\
\\
Also, be aware that there are more penalties that are recommended to be used for good performance. However, they will be explained later on the penalty optimization part (since they are not essential to obtain a valid path).

\subsubsection{Normalization}

To obtain good performance, it is important to normalize the \(Q\) matrix. Large weights and possibly inconsistent penalties can confuse the solver. The terms with high weight would likely dominate, and the other penalties would get overshadowed, so they are not treated as they should. In general, you should always perform normalization, which makes the answer more consistent and even faster (calculations with small numbers seem to help reduce time). \\
\\
However, there are different normalization scales. In mathematics, the standard normalization is usually the -1,+1 unit vector normalization. However, other scales may perform better when formulating QUBO. Not only do they perform better depending on the normalization scale, but different solvers benefit from different scales, meaning there is no unique global value to set for the normalization. Specifically, after testing different scales, I realized that the 2.0 normalization scale worked better in small-number variables scenarios for QAOA (Pennylane), and the 1.0 normalization scale worked best in all other scenarios. In my experience, 3x3 grids performed better with a 2.0 scale, while 5x5 grids worked better with a 1.0 scale. Therefore, we can assume that anything less than around 60-100 variables can do well with a 2.0 scale, while anything higher than 100+ variables should perform better with a 1.0 scale. The case for the D-Wave annealer seems different, although at first one may see that it can follow the same scale as QAOA, it actually benefits from setting high normalization scales like 5.0; by doing this, you can reduce the number of reads and get results faster while keeping or even improving accuracy.\\
\\
It may seem irrelevant, but you can generalize penalties weight independent of problem size by only changing normalization scales. Which is a huge advantage. Now you would not need to focus on choosing the correct weights, but on choosing the correct normalization (only 1 number). And further, we will see that with the time horizon approach, you will only have to worry about choosing timesteps.

\subsubsection{Heuristics}

Heuristics are recommended to improve the performance of the QUBO. In the grid-map scenarios, \textbf{Manhattan Distance} becomes interesting. It will help define a good approximation of how many time steps one should set to solve the path planning problem; it can also be used to add extra penalties that facilitate the solver.\\
\\
Manhattan distance is a way to measure distance between two points on a grid, named after the street layout of Manhattan, New York. Instead of measuring the straight-line distance (Euclidean), Manhattan distance calculates the distance you would travel if you could only move horizontally and vertically - like walking through city blocks where you cannot cut diagonally through buildings.
\[
\text{Manhattan Distance } = |x_2-x_1|+|y_2-y_1|
\]
Manhattan distance represents the minimum number of moves needed to reach your destination when you can only move in four directions (up, down, left, right). It is the theoretical best-case scenario, assuming there are no obstacles in the middle.\\
\\
For graphs, we use the \textbf{Euclidean distance}, which is the way to measure the straight line between two nodes (which would normally be the smallest distance).
\[\text{Euclidean Distance}=\sqrt{(x-x_{goal})^2 +(y-y_{goal})^2}\]

By considering this extra information, we can optimize the QUBO implementation. For example, we know that any answer that reaches the goal requires at least the same or more timesteps than the Manhattan distance between the start and the goal; otherwise, that answer is invalid. The agent could also be encouraged to follow directions that decrease the Manhattan distance to approximate the optimal answer. 

\subsubsection{Penalties Optimization}

We already saw the base penalties that make the QUBO suitable for path planning. Now, we will focus on adapted penalties that will improve the performance of the QUBO both in time and accuracy.\\
\\
\title{\textbf{Improved Goal Penalty}}
\\\\
First, we will modify the end condition penalty to a more suitable \textbf{goal penalty}. Ideally, we do not apply the goal penalty only to the last-time step, because this would only work if we know the exact number of time steps needed beforehand (which usually we do not know), else you are not encoding for the optimal solution, rather you would be constraining by making the robot think it should only reach the goal at last step. Since we are considering scenarios where staying on the same cell is penalized, enforcing a specific cell requires a specific sequence, meaning that the robot will need to stay around the cell until it can satisfy it. Secondly, we will consider that the goal penalties cannot be applied at time 0 because then it would conflict with the initial condition penalty\\
\\
For this goal penalty, one may think of 3 different approaches: \textbf{constant-time approach}, \textbf{early-time approach}, and \textbf{late-time approach}.\\
\\
In \textbf{constant-time} approach, you apply the same penalty to all time steps. Initial experiments seemed to work very well, but the problem with this approach is that it inherently does not treat teleportation, so the solver sometimes finds a local minima where it teleports directly to the goal. To make it work, you need to finely tune the goal and adjacency constraint, but it does not scale too well, eventually leading to teleportation problems or directly not finding the goal. Moreover, when including obstacles, goal and adjacency penalties do not mix that well.\\
\\
In the \textbf{early time} approach, you apply higher penalties at early steps; in other words, it encourages reaching the goal "early". The problem is that it enforces too much teleportation, making it jump to the goal at 2nd step, or at later steps, it just walks around the goal but never reaches it. Of all three, it is the most difficult to tune and achieve a high success rate.\\
\\
In \textbf{late-time} approach, higher penalties are applied at late steps, meaning that at every iteration the penalty increases (recommended up to 2x). The results show this to be the most effective goal penalty. It fixes the teleportation issue at the second step and also does well in finding valid paths since it increases very high at the last steps.\\
\\
\title{\textbf{Backtracking}}
\\\\
The purpose of this penalty is to avoid backtracking. Although theoretically there is nothing wrong with going back in order to find the correct answer, it also means that there is a more direct way to reach the goal (basically, the one that does not backtrack). Therefore, it would be nice for the solver to avoid visiting the cells twice and find the path directly. The idea is to enforce the optimal solution, and a backtracking solution will never be optimal. \\
\\
It also helps with goal penalty in general, since it avoids wandering around the goal or start (which is normal behavior when it does not find the goal or when it teleports to the goal).\\
\\
This is a global constraint,not per time step, so we need to penalize any pair of time steps $t_1 < t_2$ where the agent is at the same non-goal position.

\[
P_{backtracking} = K_{bt} \cdot \sum_{(i,j)\ne(e_i,e_j)} \sum_{t_1<t_2} x_{i,j,t_1} x_{i,j,t_2}
\]

This penalty does not apply to the goal cell; you can stay/visit the goal more than once; otherwise, it would conflict with the goal lock penalty.\\
\\
\title{\textbf{Teleportation Penalty}}
\\\\
The idea is to prevent early arrival (hereafter referred to as \textbf{teleportation}) by adding a time-based heuristic penalty. This is easily done by penalizing reaching the goal before it is physically possible (taking Manhattan distance into account to determine the minimum number of steps needed). In practice, this penalty helps balance the goal and greatly improves the results.

\subsubsection{Time window approach}

In a time-windowed QUBO approach, you break a long-horizon  \cite{RecedingHorizon} (e.g., planning over 100 time steps) into shorter, overlapping, or sequential windows (e.g., 5–10 steps each). This approach is used because QUBOs with too many variables become unsolvable (there are not enough qubits or the solver tends to give bad solutions/ignore penalties). \\
\\
It requires a special implementation because, in a way, you want to relate each QUBO to some things, but you want them to be implemented independently for others. Ideally, you keep a history of what happened in the past and have an awareness of its purpose. The first thing we need to know is that they are implemented sequentially in incremental order, meaning that it is not possible to have a third time-window QUBO without the second, and so on (it will not make QUBO parallelizable).\\
\\
The first thing to take into account for implementing the time-window approach is that you need to treat each QUBO essentially independently. Each window is a new, clean QUBO with a reset index; this way, each window is easier to code and understand. From previous QUBOs, you need only one thing: the path. With the path, you must ensure that the start of one QUBO matches the end of the previous QUBO. It is necessary to maintain path continuity. The problem with this is that it is logically easy to understand, but a bit difficult to code because you need to handle that mix. You can decide to keep twice and have something like ([2,2,5],[2,2,6]), or you would like to mix them like it was solved in a single QUBO. \\
\\
The best scenario is the mixing, because the user at the end does not want to know how it got the answer, but to get a uniform/consistent answer every time (independent of solution method). To do that, the solver behind the scenes needs to make sure position matches and readjust the time steps; otherwise, there will be a jump in time (something like ([3, 4, 4],[4, 4, 6], where time 5 is skipped because of the merge. In addition, to obtain better results, you not only use the last position but also add the whole path to the visited list, so those cells will be affected by backtracking in future windows, but they are softened in case some window did not perform as expected. \\
\\
It may seem difficult to implement at first, but its advantages outweigh the difficulties. With the time-window approach, you can virtually solve any path-finding problem independent of size. You just need to separate a big problem into many time steps that will be assembled into a solution. This solves the issue of the current technology of quantum computers. It enables the possibility of solving large problems even with a small number of qubits at a time; basically, it is necessary to face quantum computer scaling issues. It can also be used for dynamic planning. When creating a new QUBO, you can change the obstacle position (add obstacles, delete obstacles), and the solver will recalculate accordingly. You can update the environment by changing conditions and weight on penalties (based on traffic, battery, and time priority).\\
\\
The problem with this method is that you need to start thinking with an approximation approach. Naturally, the solver will not reach the goal until the very last QUBO, so you do not aim for the solution in all the QUBOs, but aim to approximate it as much as you want (so as to facilitate following QUBOs until one is able to reach the goal).\\
\\
\title{\textbf{Approximation Penalty}}\\
\\
This is a modification of the goal penalty for the time-window approach. As previously stated, the only window that will reach the goal is the last one; that is, most windows will not be able to satisfy the goal penalty. To solve this problem, the weight of the goal on the initial windows is reduced to avoid teleportation scenarios. Because the whole idea changes from reaching the goal to getting to a better position closer to the goal, I use a heuristic approach that reduces the Manhattan/Euclidean distance. Therefore, at the start, the solver checks if the current position is physically reachable in this window, that is, it verifies that \(\text{distance to goal } < \text{Window Total Time}\); if it is not reachable, the goal penalty is changed to the approximation penalty. However,  this weight needs to be high enough to approach the goal but small enough not to override other penalties, especially backtracking.\\
\\
In mazes or scenarios with a high obstacle density, it is possible that to reach the goal, one must first deviate from the goal; that is, the heuristic distance will not be minimized. Path-planning in this scenarios is mainly solved by two penalties, the first penalty was previously showed and it is \textbf{backtracking}, it avoids that windows collide between them, that is, it avoids re-writing previous path, the windows will detect that reducing distance was tried by previous windows and did not work, so it decides to continue path as previous windows intended, essentially, if you are coming from a corridor the approximation penalty will not be high enough so new windows not backtrack the corridor or previous path. The second idea is to compute an obstacle potential field for grid maps and a node connectivity potential for graph maps. With this, we not only reduce the distance but also encourage the solver to move to open areas where the obstacle density is smaller, and the solver has more freedom to move. This is also useful for multi-robots, where they should not be cluttered in small areas.

\subsection{Pre-Processing}

After applying optimization techniques, heuristics, and the time horizon approach, there is not much that can be done to improve the performance in terms of time or accuracy. Therefore, pre-processing is a necessary system-level optimization. Pre-processing plays a crucial role in QUBO, and there are many mathematical proof methods to boost performance with variable fixing via algorithms \cite{Pre-process1} \cite{Pre-process2}.\\
\\
After thoroughly trying these mathematical methods for reducing a QUBO, unfortunately, path planning does not present numeric characteristics that allow for this direct mathematical method. However, we know by logic that we can consistently reduce more than half of the variables needed for the QUBO representation.  For example,for the first iteration, the start position should always be set to 1 (active), and the rest should be set to 0. It is also obvious that all obstacle variables can be set to 0 at all time steps.\\
\\
Then, there are some variables that one could naturally guess should be set to 0; however, they are not that straightforward. For example, based on adjacency, at time step 2, you could have at most four possible good bits, which are related to the connectivity of the problem (up, down, left, right), and you could go on forward. To determine which positions are reachable, a breadth-first search (BFS) algorithm is applied. We want to reduce our complete state space to the search tree-reduced space. This way, instead of saving all variables, you would only need to save something like $$ 0: (2, 0), 1: (1, 0), (2, 1), 2: (2, 2), (0, 0) $$

where the time step and the visitable positions are saved at that time step. This pre-processing alone plays a huge role in performance; the BFS can be configured to not consider itself or already visited positions as possible options, which will significantly reduce the number of variables. \\
\\
In fact, this BFS pre-processing is the biggest optimization in all the code, and this alone makes the code faster and more consistent than any other approach. However, we also need to learn how to exploit pre-processing performance. At each time step, the space reduction becomes less and less; this means that after a certain point, one does not want to waste time on continuing with the space reduction. Therefore, if you mix this with a time windowing technique, better results can be obtained than solving the complete QUBO at once (even if you have enough qubits to do so). Because each new QUBO window is considered a fresh start; thus, the BFS is recalculated, resulting in a significant reduction from the initial steps. By separating into multiple QUBOs, the total number of variables used would be reduced, rather than if all are computed at once, just from the aggressive nature of how the most powerful pre-processing techniques work. One can realize that the idea of pre-processing is to exploit the job of the penalties; if you need to define it as a penalty, you will also probably be able to apply pre-processing to it.\\
\\
Theoretically, based on the off-diagonal and diagonal coefficients, one can calculate the best and worst scenarios, and hence reduce the variables to 0 or 1. However, this QUBO type is so logically constrained that it will never be reduced based on this rule alone, given that penalties are always too coupled. Therefore, we are following a more logical approach; however, this logical reduction approach has a limit, and at some point, you have to make a choice that cannot possibly be reduced beforehand. The more accurate numerical method is to compare only the diagonal elements; this way, you can be more certain of the good states. However, it only works (safely) to set 0 the diagonals that are too off, which are not that many (maybe one per time step). You could also try setting the diagonal elements that are too high compared to the rest to 1, but it is not safe (consider that the QUBO may benefit cells that solely have more connections, but this is not always a good path). The interesting thing about the numerical approach is that it does not prune many answers, but it can be done iteratively, that is, clearing at one iteration updates the coefficients and allows you to continue fixing variables on newer iterations.\\
\\
By doing this, the pre-processing becomes so aggressive that it may obtain the optimal path just by pre-processing, without even the need to call for a solver. This is the perfect scenario (fastest response and optimal answer). However, the diagonal numerical approach seems a bit problematic in cases where the heuristic does not match (i.e., if reducing the Manhattan distance is the heuristic, but in reality is not the best approach, it will conflict with the pre-processor). Sometimes, the solution does not reduce the total distance because there could be obstacles in between. Therefore, if you implement it badly, the pre-processor will not know that it is a bad choice, and the QUBO will never work. This is not a fault in the numerical approach but a fault in the QUBO building.

\subsection{Multi-Agent}

To date, the focus has been on single-agent scenarios, where the robot is free to choose any path as long as it follows the penalties (adjacency, one-hotness, and no collision). However, standardized and super-tested algorithms for classical path planning, such as Dijkstra or A*, already exist. The area where QUBO formulation shines is in multi-agent scenarios; in fact, it can even benefit from that and perform faster computations. \\
\\
Therefore, I decided to follow a different approach that I believe should be the rule and not the exception when designing navigation frameworks in the future. Path planning should be treated as a multi-agent system by nature, and single-path planning should be treated as a degenerate case where n = 1. Current navigation and planning software focuses only on single-scenario cases, and although they support multi-robot at a small scale, they are not suitable for larger systems. I hypothesize that the cause is the lack of algorithms, but QUBO is a promising alternative for addressing these limitations. With the QUBO formulation, not only can you reduce resource usage, save one map for all robots, and have total control of each robot, but also have a linear growth in variable count with respect to the number of agents. In other words, in QUBO, there is inherently no difference between solving for single or multiple agent environments; at the end of the day, the increase in variables is linear, and all robots use approximately the same number of variables (depending on active time steps). With pre-processing, having multiple robots may become an advantage rather than a limitation; more variables can be reduced by constraining the general map space. Therefore, there can be scenarios in which more robots imply less total variable usage, and hence, faster computation. However, to perform multi-agent in QUBO, we need to do two major adaptations, \textbf{time modification} and \textbf{crash avoidance}.

\subsubsection{Time}

Thus far, time implementation has not been an issue because for a single robot, you only need to start at time 0 and maintain continuity. However, when handling multiple robots, there is no guarantee that they will start at the same time step or remain active for the same duration.\\
\\
To implement the timeline, we will follow a global clock approach, which is a centralized approach that will require all robots to connect to the same solver. Although distributed systems are generally preferred in robotics, the global clock is easier to implement than clock synchronization between robots.\\
\\
We also need to modify a bit how the indexing will work, since with a global clock, we will have global time. We define the end to be the time of the last active robot. If that robot is expected to be active until time T, then to keep a fairly linear index, we say enumerate variables to be MxNxT for robot 1, then MxNx2T for robot2 and so on. This maintains coherent linear indexing, and sparse algebra solvers efficiently handle the unused variable spaces (treating absent variables as 0). Since most robots won't use their full allocated variable space, this approach remains computationally efficient.\\
\\
Then, to maintain a logical time order, time 0 will be related to the first group of active robots (could be only one), and all other robot starts will be relative to it, meaning that one robot may have a local time from 0 to 6 but have a relative start of 2, meaning that their time in the global clock would be managed as 2-8. This relative timing makes collision penalties straightforward to implement, as they only need to be applied between robots that are active simultaneously.

\subsection{Post-Processing}

As stated in the pre-processing, the idea is to exploit the job of penalties; if a penalty is defined, it can either be used for pre- or post-processing.\\
\\
Post-processing converts erroneous or suboptimal answers into optimal paths. Initially, we attempted to solve two common problems: the first is related to one-hotness and multiple active cells at the same time step, and the second is teleportation issues.\\ 
\\
\title{\textbf{Continuity check}}\\
\\
The idea is to solve one-hotness errors,i.e, scenarios in which multiple active cells exist in the same time step. This is a very common error because the K\_hot penalty and the K\_adj penalty disrupt each other because they are highly related (at least mathematically sensible enough for one to overshadow the other). Usually, when this problem occurs, it is because the solver was not able to decide between two routes; however, at the next step, that does not happen, as it is meant to always follow a route, but at some time step, it could not decide between two options; therefore, it selects both. These scenarios can be easily solved by selecting the cell that maintains continuity and dropping all the others.\\
\\
\title{\textbf{Invalid Move}}
\\\\
Sometimes, the robot should reach the goal but deviates a bit; when it tries to go back to the goal, it cannot because of the backtracking penalty, so it teleports to the goal. There are different approaches to solve this, but for now, we employ the simplest of all, that is, restarting the window. In general, if the original solver has high accuracy by repeating bad windows, it should fix all bad path issues. However, this area allows for more thorough experimentation, such as resetting from the last good cell, i.e., the one before the invalid move, or maybe a range before. This could slightly reduce time by not reconstructing the whole window but advancing a bit, basically as if the window was meant from the beginning to be size Error\_T - 1; however, it is not as easy to determine, and improvements do not seem that critical since resetting the window gives good results.\\
\\
\title{\textbf{Clash conflict solver}}
\\\\
Typically, modern algorithms treat each robot path as independent and then solve for crash-disruption points. This idea can also be translated into QUBO, either going for the optimal answer and re-planning by adjusting/encouraging some directions depending on the results, or making the robot wait enough in some cell of its path to avoid the collision, and then continuing the path as intended. Although this will probably never be an optimal solution, it is a fast/good fallback when the answer suffers from collision.

\section{Benchmark and Results}

Benchmark results on current hardware indicate that QUBO-based approaches do not yet outperform classical path planners for the scenarios tested. For single-robot cases, classical algorithms are orders of magnitude faster (\(0.001s\) vs \(0.003s\)). However, the performance gap narrows in multi-agent scenarios: on 10×10 grids with 4 robots, QUBO achieves near-optimal solutions (5.3\% path length penalty) with computational times that, while still slower (\(0.563s\) vs \(0.003s\)), demonstrate more favorable scaling properties than classical sequential planning. \\
\\
The current limitations stem from (1) classical simulation overhead, (2) limited qubit counts on available quantum hardware, and (3) QUBO matrix size constraints for high-resolution maps. These results establish a baseline for future quantum hardware where the theoretical linear scaling advantage may materialize.\\
\\
On current scales (10×10), classical MAPF is more efficient. The QUBO advantage may emerge at larger scales, where the classical search space becomes prohibitive, and quantum hardware may be beneficial. We established a baseline on 5×5 and 10×10 grids. The variable explosion QUBO (O(M×N×T×R)) limits testing on higher scales, motivating future research on decomposition methods. In addition, at these scales, simulated annealing or simulated QAOA performs better than real Qiskit hardware (did not try with D-Wave computers), but at high qubit numbers, the quantum computer could not handle it well. Ideally, with a high number of variables and optimization for quantum hardware, the algorithm should outperform the classical algorithm.\\
\\
The purpose of these experiments is not to demonstrate superiority over classical MAPF, but to establish a baseline, characterize failure modes, and evaluate scaling behavior under increasing agent density.\\
\\
\begin{figure}[htbp]
\centering
\begin{minipage}{0.45\textwidth}
  \centering
  \includegraphics[width=\linewidth]{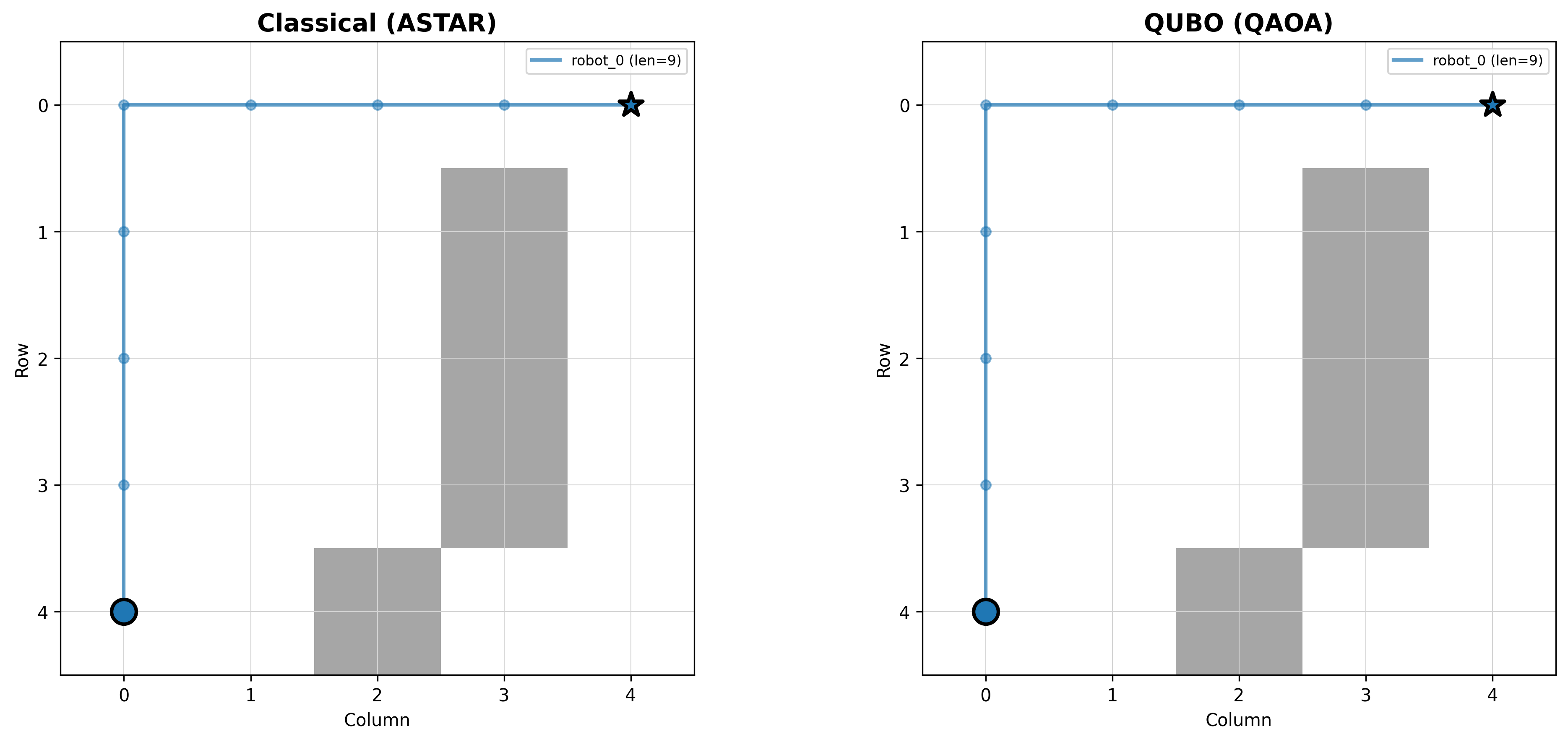}
  \caption{A* (left) and QUBO solved with simulated QAOA (right) on a 5x5 grid}
\end{minipage}\hfill
\begin{minipage}{0.45\textwidth}
  \centering
  \includegraphics[width=\linewidth]{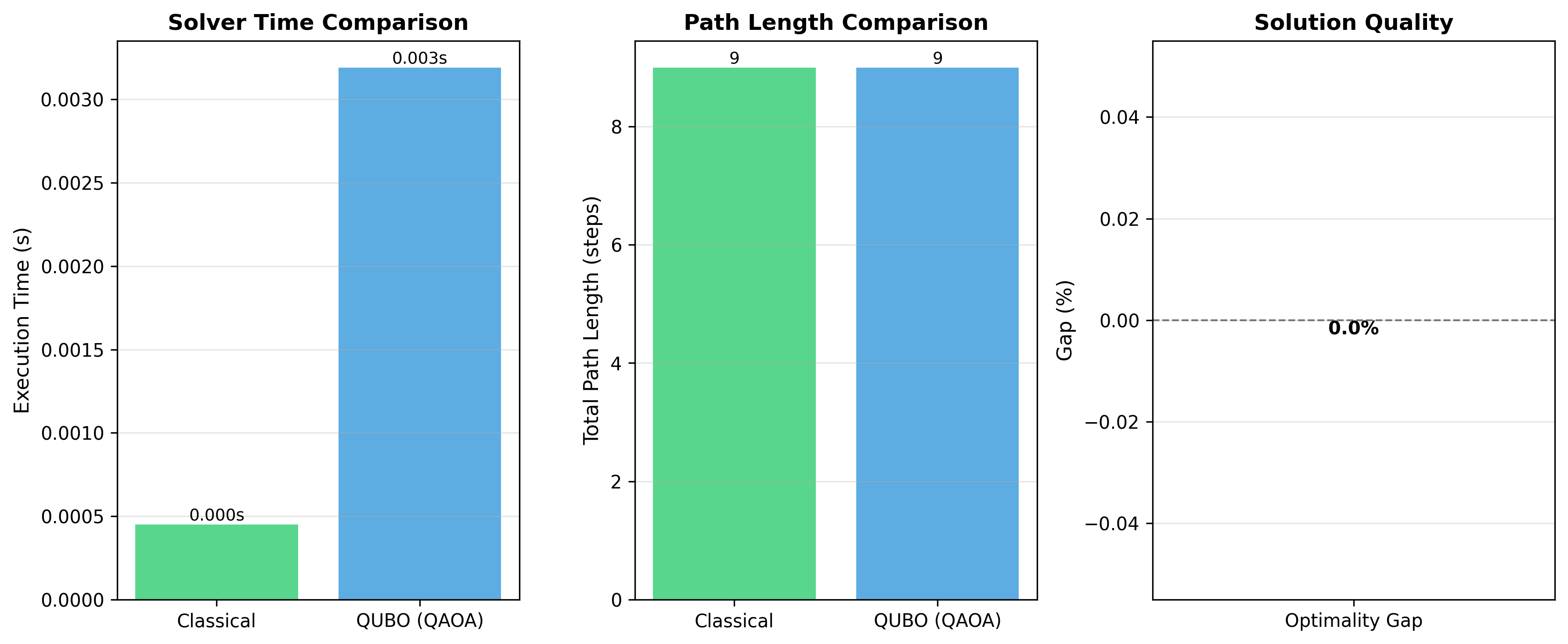}
  \caption{QUBO and Classical get to the same optimal answer. Dijkstra is faster}
\end{minipage}
\end{figure}

\begin{figure}[htbp]
\centering
\begin{minipage}{0.45\textwidth}
  \centering
  \includegraphics[width=\linewidth]{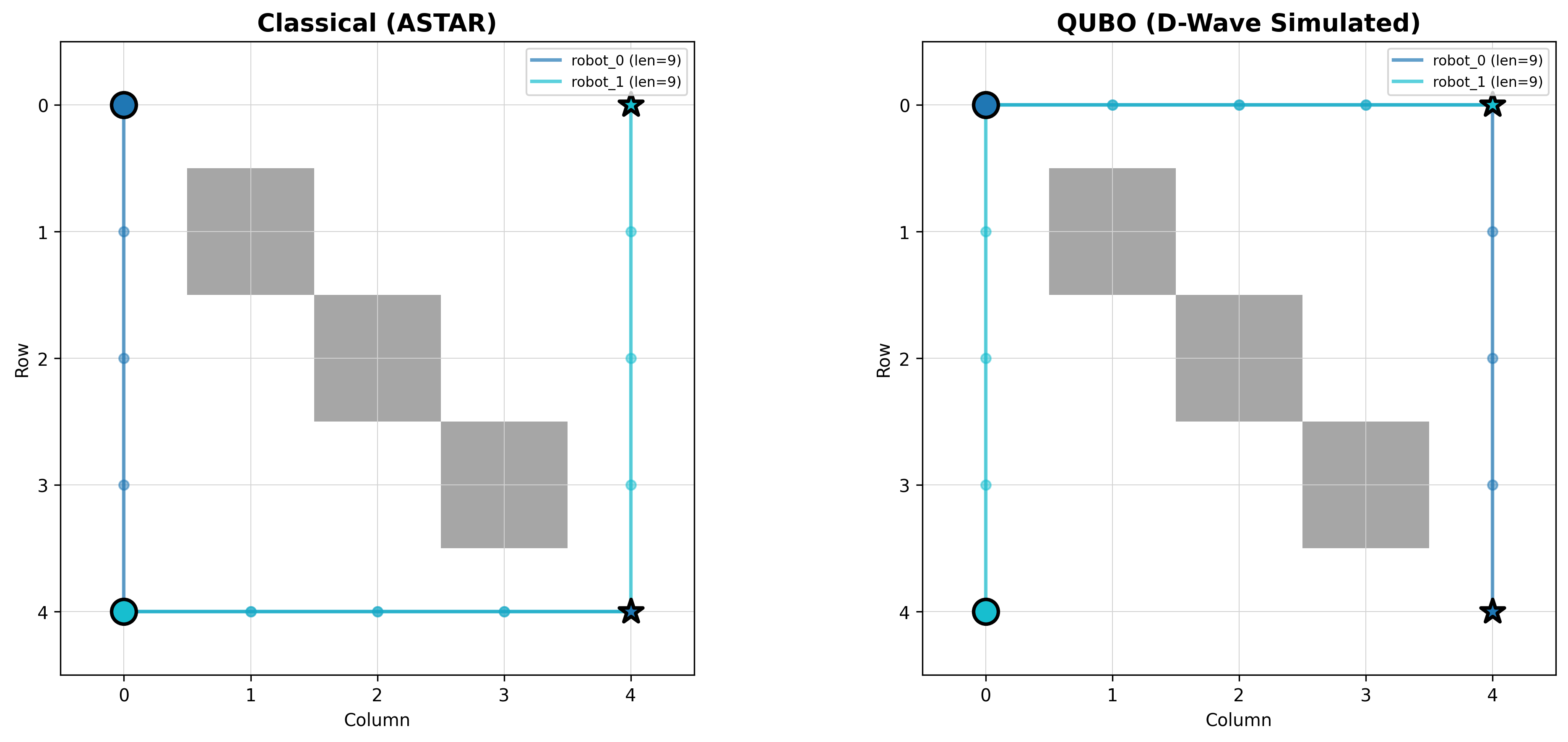}
  \caption{A* (left) and QUBO solved with simulated annealing from D-Wave (right) on a 5x5 grid with obstacle in the middle and two robots}
\end{minipage}\hfill
\begin{minipage}{0.45\textwidth}
  \centering
  \includegraphics[width=\linewidth]{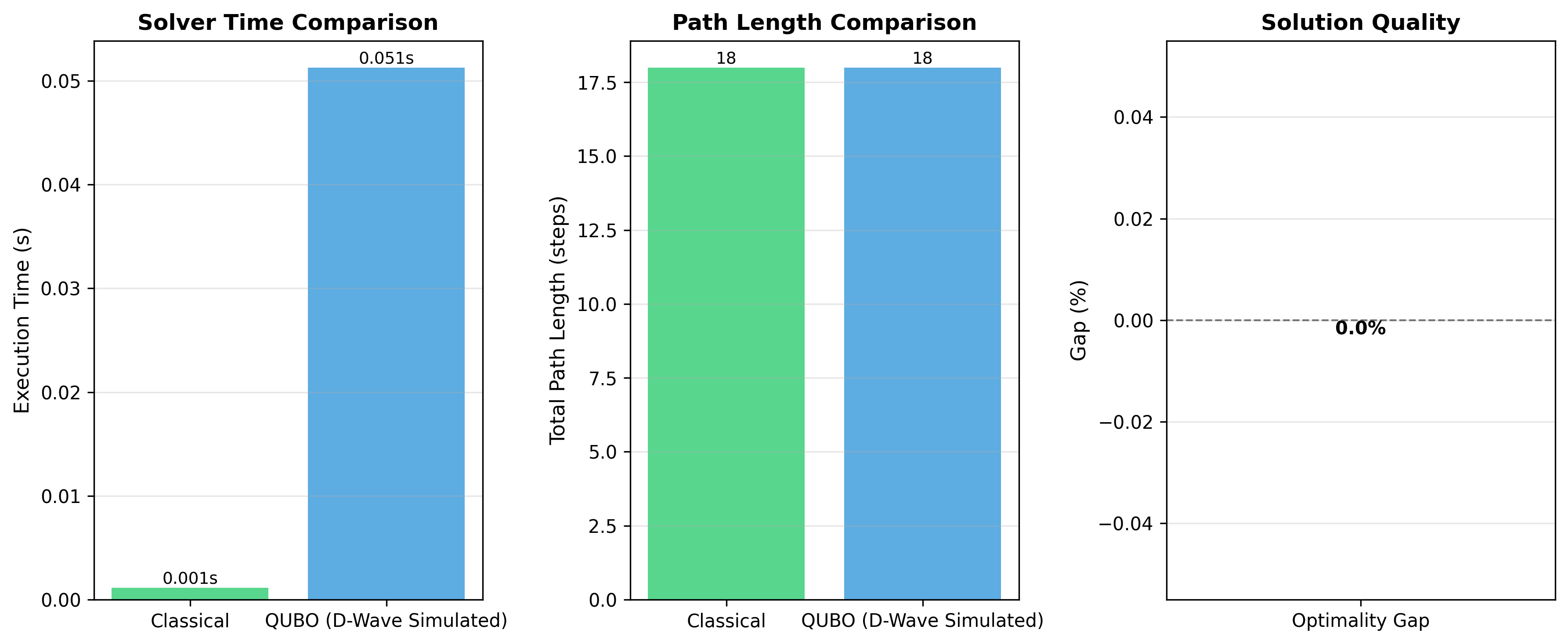}
   \caption{Both get the same optimal length of 18 steps, but QUBO chooses a different orientation (generally, when multiple optimal paths are chosen at random by the solver)}
\end{minipage}
\end{figure}

\begin{figure}[htbp]
\centering
\begin{minipage}{0.45\textwidth}
  \centering
  \includegraphics[width=\linewidth]{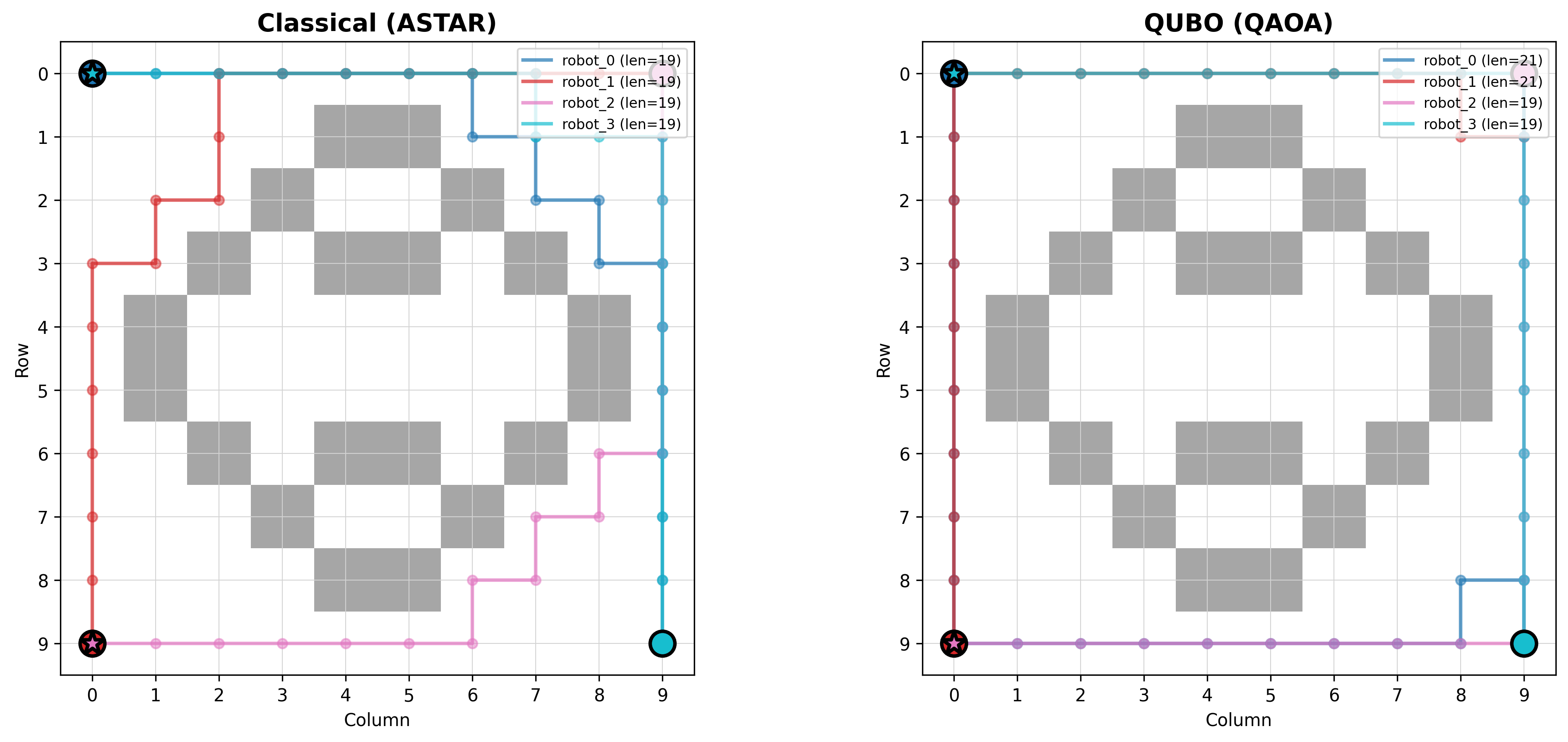}
  \caption{A* (left) and QUBO solved with simulated QAOA on a 10x10 with high obstacle density and 4 robots in each corner}
\end{minipage}\hfill
\begin{minipage}{0.45\textwidth}
  \centering
  \includegraphics[width=\linewidth]{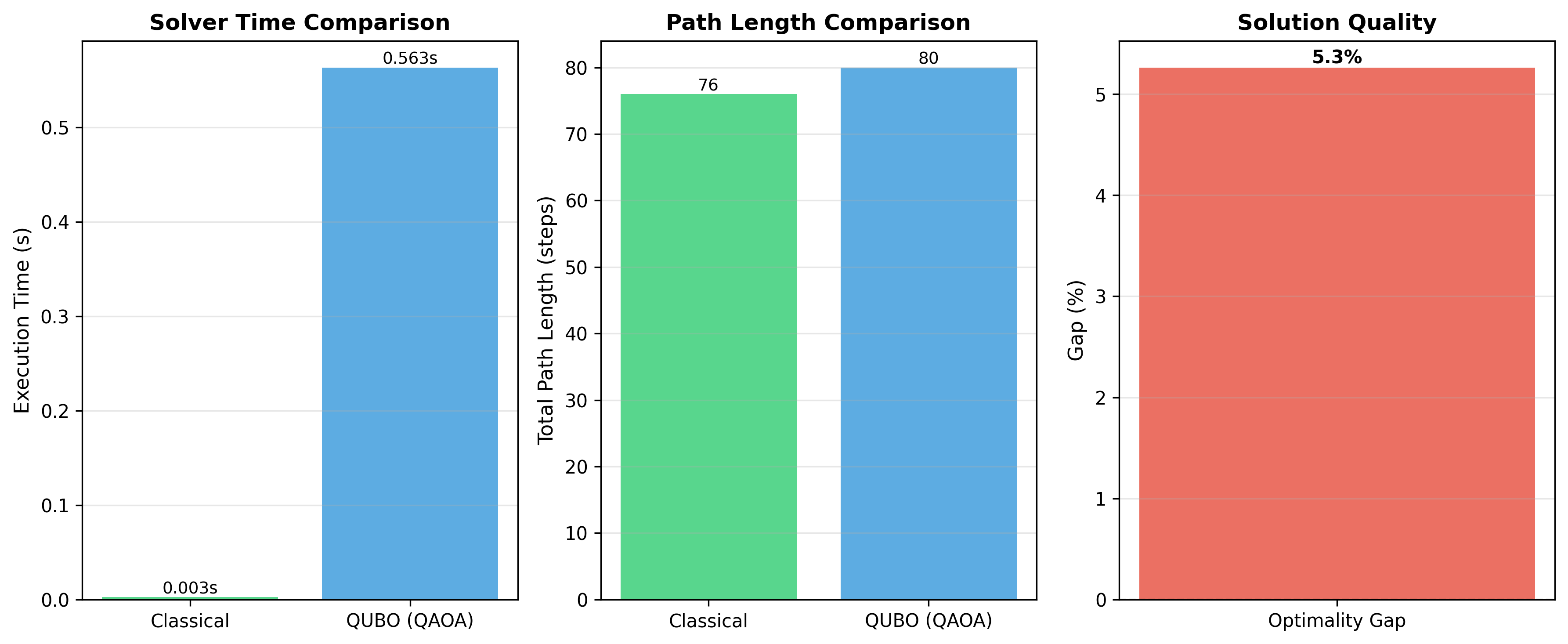}
  \caption{QUBO is generally optimal, but sometimes it stays around the goal (local minima) and slightly increase number of steps}
\end{minipage}
\end{figure}

\FloatBarrier
\begin{table}[htbp]
    \centering
    \caption{Performance comparison of Classical MAPF vs QUBO-based approach}
    \label{tab:benchmark_results}
    \begin{tabularx}{\textwidth}{lccccc>{\centering\arraybackslash}X>{\centering\arraybackslash}X}
    \hline
    \textbf{Problem} & \textbf{Grid} & \textbf{Robots} & \textbf{Classical} & \textbf{QUBO} & \textbf{Path (C/Q)} & \textbf{Variables} & \textbf{Reduction} \\
     & & & \textbf{Time (s)} & \textbf{Time (s)} & \textbf{(steps)} & \textbf{(orig/red)} & \textbf{(\%)} \\
    \hline
    5×5 Single  & 5×5   & 1 & $<$0.001 & 0.003 & 9/9    & 300 $\to$ 12 & 96\%\\
    5×5 Multi   & 5×5   & 2 & 0.001  & 0.051 & 18/18  & 500 $\to$ 16 & 96.8\%\\
    10×10 Multi & 10×10 & 2 & 0.003  & 0.563 & 76/80  & 9600 $\to$ 0 & 100\%\\
    \hline
    \end{tabularx}
    {\footnotesize
    Classical uses A* with prioritized planning. QUBO solved using D-Wave simulated annealing and simulated QAOA (PennyLane implementation).
    }
\end{table}

\FloatBarrier
Observe that QUBO behaves differently because it was programmed to avoid dense routes when using the time-windowed approach. Since the program does not know whether it will reach a dead end. By analyzing the table \ref{tab:benchmark_results}, one can get an idea of the importance of pre-processing for QUBO solving and how powerful it is. For a fast scenario, you always want a 95+\% reduction or full pre-processing (which is generally achieved with the techniques given, that is, BFS + Time windowing).

\FloatBarrier
\section{Conclusions and Future Work}

Scalable coordination in multi-robot systems remains a central challenge as the number of agents and environmental complexity increase. A QUBO-based formulation is presented for scalable multi-robot path planning, positioning combinatorial optimization as an alternative to classical sequential MAPF approaches. By encoding multi-agent coordination within a unified quadratic objective, the formulation achieves linear growth in decision variables with respect to the number of agents.\\
\\
To address practical constraints, we introduced BFS-based logical pre-processing -achieving over 95\% variable reduction-, adaptive penalty strategies for enforcing robotic constraints, and a time-windowed decomposition framework that enables execution under current hardware limitations. Experimental results on 5×5 and 10×10 grids with up to four robots demonstrate near-optimal performance while highlighting current computational trade-offs relative to classical planners.\\
\\
Future work will explore distributed QUBO formulations for truly decentralized multi-robot systems, binary encoding to reduce qubit requirements, and integration with Variational Quantum Eigensolvers (VQE) for improved solution quality. As quantum hardware continues to mature, the linear scaling properties of QUBO-based multi-agent planning may realize practical advantages over classical exponential methods in large-scale robotic deployments.\\
\\
This work establishes a practical baseline for quantum-ready multi-robot coordination and provides an open foundation for continued exploration at the intersection of robotics and quantum optimization. Although current quantum hardware does not yet provide computational advantage at the tested scales, the proposed formulation establishes a structured baseline for future hardware improvements and hybrid quantum-classical solvers.\\
\\

You can check all the code at \url{https://github.com/JavideuS/Spooky}

\section{Limitations}

The best and worst aspects of this algorithm are its high dependency on pre-processing. For a higher reduction, the pre-processing sometimes makes a guess, which may lead to a faulty situation or an increase in time. The second limitation is the crash penalty. Although crash avoidance is simple and already implemented, swap-collision is not quadratic by nature but quartic, which requires either a new efficient encoding/logic to be implemented or a considerable increase in the QUBO variables. The third major limitation is the centralized nature of the protocol; the entire planning process can be completed by a single solver. The problem is that robotics is meant to be distributed, and current protocols/systems are serverless by preference; ultimately, a single-point failure is a significant disadvantage. Finally, the solution is not guaranteed to be found; if well formulated, it should be theoretically complete based on the assumption of the adiabatic theorem; however, in real life, this is not always the case. Although implementations can and should be made to make it complete (better encoding reduces randomness). The most obvious solution is that QUBO is solved ideally by quantum computers; however, the problem is that there is an ongoing shortage of functional qubits. If a high number of qubits are used today (approximately 100-150 qubits), there are high probabilities of noise, which cannot be solved directly by the QUBO and may require implementing redundancy to make it noise-proof.

\nocite{SparseLinearAlgebra}
\nocite{QuantumAnnealer}

\bibliographystyle{unsrt}
\bibliography{references} 

@misc{Pre-process1,
      title={Quadratic Unconstrained Binary Optimization Problem Preprocessing: Theory and Empirical Analysis}, 
      author={Mark Lewis and Fred Glover},
      year={2017},
      eprint={1705.09844},
      archivePrefix={arXiv},
      primaryClass={cs.AI},
      url={https://arxiv.org/abs/1705.09844}, 
}

@misc{Pre-process2,
      title={Logical and Inequality Implications for Reducing the Size and Complexity of Quadratic Unconstrained Binary Optimization Problems}, 
      author={Fred Glover and Mark Lewis and Gary Kochenberger},
      year={2017},
      eprint={1705.09545},
      archivePrefix={arXiv},
      primaryClass={cs.AI},
      url={https://arxiv.org/abs/1705.09545}, 
}

@misc{SparseLinearAlgebra,
  title        = {Sparse Linear Algebra},
  author       = {Tim Mattson and Mark Murphy},
  url = {https://patterns.eecs.berkeley.edu/?page_id=202},
}

@misc{QuantumAnnealer,
	title = {Quantum Annealer},
    author = {QuEra},
	url = {https://www.quera.com/glossary/quantum-annealer},
}

@misc{DwaveQA,
	title = {What is Quantum Annealing?},
    author = {D\-wave Quantum},
	url = {https://docs.dwavequantum.com/en/latest/quantum_research/quantum_annealing_intro.html},
}

@article{Kadowaki_1998,
   title={Quantum annealing in the transverse Ising model},
   volume={58},
   ISSN={1095-3787},
   url={http://dx.doi.org/10.1103/PhysRevE.58.5355},
   DOI={10.1103/physreve.58.5355},
   number={5},
   journal={Physical Review E},
   publisher={American Physical Society (APS)},
   author={Kadowaki, Tadashi and Nishimori, Hidetoshi},
   year={1998},
   month=nov, pages={5355–5363} 
}

@misc{farhi2014quantumapproximateoptimizationalgorithm,
      title={A Quantum Approximate Optimization Algorithm}, 
      author={Edward Farhi and Jeffrey Goldstone and Sam Gutmann},
      year={2014},
      eprint={1411.4028},
      archivePrefix={arXiv},
      primaryClass={quant-ph},
      url={https://arxiv.org/abs/1411.4028}, 
}

@misc{QAOA,
	title = {{What is the QAOA?}},
    author = {OpenQAOA},
	url = {https://openqaoa.entropicalabs.com/what-is-the-qaoa/},
}

@misc{mapf,
author = {Stern, Roni},
year = {2019},
month = {10},
pages = {96-115},
title = {Multi-Agent Path Finding – An Overview},
isbn = {978-3-030-33273-0},
doi = {10.1007/978-3-030-33274-7_6}
}

@ARTICLE{RecedingHorizon,
  author={Zhang, Bin and Tang, Liang and DeCastro, Jonathan and Roemer, Michael J. and Goebel, Kai},
  journal={IEEE Transactions on Industrial Electronics}, 
  title={A Recursive Receding Horizon Planning for Unmanned Vehicles}, 
  year={2015},
  volume={62},
  number={5},
  pages={2912-2920},
  doi={10.1109/TIE.2014.2363632}
}

@article{SHARON2015_CBS,
title = {Conflict-based search for optimal multi-agent pathfinding},
journal = {Artificial Intelligence},
volume = {219},
pages = {40-66},
year = {2015},
issn = {0004-3702},
doi = {https://doi.org/10.1016/j.artint.2014.11.006},
url = {https://www.sciencedirect.com/science/article/pii/S0004370214001386},
author = {Guni Sharon and Roni Stern and Ariel Felner and Nathan R. Sturtevant},
}

@misc{cap2014prioritizedplanningalgorithmstrajectory,
      title={Prioritized Planning Algorithms for Trajectory Coordination of Multiple Mobile Robots}, 
      author={Michal Čáp and Peter Novák and Alexander Kleiner and Martin Selecký},
      year={2014},
      eprint={1409.2399},
      archivePrefix={arXiv},
      primaryClass={cs.RO},
      url={https://arxiv.org/abs/1409.2399}, 
}

@article{lucas2014ising,
   title={Ising formulations of many NP problems},
   volume={2},
   ISSN={2296-424X},
   url={http://dx.doi.org/10.3389/fphy.2014.00005},
   DOI={10.3389/fphy.2014.00005},
   journal={Frontiers in Physics},
   publisher={Frontiers Media SA},
   author={Lucas, Andrew},
   year={2014} 
}

@misc{neukart2017traffic,
      title={Traffic flow optimization using a quantum annealer}, 
      author={Florian Neukart and Gabriele Compostella and Christian Seidel and David von Dollen and Sheir Yarkoni and Bob Parney},
      year={2017},
      eprint={1708.01625},
      archivePrefix={arXiv},
      primaryClass={quant-ph},
      url={https://arxiv.org/abs/1708.01625}, 
}

@misc{venturelli2016job,
      title={Quantum Annealing Implementation of Job-Shop Scheduling}, 
      author={Davide Venturelli and Dominic J. J. Marchand and Galo Rojo},
      year={2016},
      eprint={1506.08479},
      archivePrefix={arXiv},
      primaryClass={quant-ph},
      url={https://arxiv.org/abs/1506.08479}, 
}

@article{Feld_2019,
   title={A Hybrid Solution Method for the Capacitated Vehicle Routing Problem Using a Quantum Annealer},
   volume={6},
   ISSN={2297-198X},
   url={http://dx.doi.org/10.3389/fict.2019.00013},
   DOI={10.3389/fict.2019.00013},
   journal={Frontiers in ICT},
   publisher={Frontiers Media SA},
   author={Feld, Sebastian and Roch, Christoph and Gabor, Thomas and Seidel, Christian and Neukart, Florian and Galter, Isabella and Mauerer, Wolfgang and Linnhoff-Popien, Claudia},
   year={2019},
   month=jun 
}

@article{Yu_LaValle_2013, 
title={Structure and Intractability of Optimal Multi-Robot Path Planning on Graphs}, 
volume={27}, 
url={https://ojs.aaai.org/index.php/AAAI/article/view/8541}, 
DOI={10.1609/aaai.v27i1.8541}, 
number={1}, 
journal={Proceedings of the AAAI Conference on Artificial Intelligence}, 
author={Yu, Jingjin and LaValle, Steven},
year={2013},
month={Jun.},
pages={1443-1449}
}

@article{Zhou_2020,
   title={Quantum Approximate Optimization Algorithm: Performance, Mechanism, and Implementation on Near-Term Devices},
   volume={10},
   ISSN={2160-3308},
   url={http://dx.doi.org/10.1103/PhysRevX.10.021067},
   DOI={10.1103/physrevx.10.021067},
   number={2},
   journal={Physical Review X},
   publisher={American Physical Society (APS)},
   author={Zhou, Leo and Wang, Sheng-Tao and Choi, Soonwon and Pichler, Hannes and Lukin, Mikhail D.},
   year={2020},
   month=jun }

@misc{rehfeldt2022fasterexactsolutionsparse,
      title={Faster exact solution of sparse MaxCut and QUBO problems}, 
      author={Daniel Rehfeldt and Thorsten Koch and Yuji Shinano},
      year={2022},
      eprint={2202.02305},
      archivePrefix={arXiv},
      primaryClass={math.OC},
      url={https://arxiv.org/abs/2202.02305}, 
}

@misc{rovara2024frameworkformulatepathfindingproblems,
      title={A Framework to Formulate Pathfinding Problems for Quantum Computing}, 
      author={Damian Rovara and Nils Quetschlich and Robert Wille},
      year={2024},
      eprint={2404.10820},
      archivePrefix={arXiv},
      primaryClass={quant-ph},
      url={https://arxiv.org/abs/2404.10820}, 
}

\end{document}